\documentclass[letterpaper]{article}
\usepackage{iccc}
\usepackage{graphicx}

\usepackage{times}
\usepackage{helvet}
\usepackage{courier}
\title{Paranoid Transformer: \\ Reading Narrative of Madness as Computational Approach to Creativity}
\author{
  Yana Agafonova \\
Higher School of Economics, \\ 
Saint-Petersburg\\\And
 Alexey Tikhonov \\
  Yandex, \\
  Berlin\\\And
   Ivan P. Yamshchikov \\
  Max Planck Institute \\
  for Mathematics in the Sciences, \\
 Leipzig\\
  {\tt ivan@yamshchikov.info}\\}
\begin{document} 
\maketitle
\begin{abstract}
\begin{quote}
This papers revisits the receptive theory in context of computational creativity. It presents a case study of a Paranoid Transformer — a fully autonomous text generation engine with raw output that could be read as the narrative of a mad digital persona without any additional human post-filtering. We describe technical details of the generative system, provide examples of output and discuss the impact of receptive theory, chance discovery and simulation of fringe mental state on the understanding of computational creativity.
\end{quote}
\end{abstract}

\section{Introduction}

The studies of computational creativity in the field of text generation commonly aim to represent a machine as a creative writer. Although text generation is broadly associated with a creative process, it is based on linguistic rationality and the common sense of the general semantics. In \cite{yamshchikov2019decomposing}, authors demonstrate that if a generative system learns a better representation for such semantics, it tends to perform better in terms of human judgment. However, since averaged opinion could hardly be a beacon for human creativity, is its' usage feasible regarding computational creativity?

The psychological perspective on human creativity tends to apply statistics and generalizing metrics to understand its object \cite{rozin2001social,yarkoni2019generalizability}, so creativity becomes introduced through particular measures, which is epistemologically suicidal for aesthetics. While both creativity and aesthetics depend on judgemental evaluation and individual taste that depends on many aspects \cite{hickman2010art,melchionne2010old}, the concept of perception has to be taken into account, when talking about computational creativity.

The variable that is often underestimated in the mere act of meaning creation is the reader herself. Although the computational principles are crucial for text generation, the importance of a reading approach to generated narratives is to be revised. What is the role of the reader in the generative computational narrative?  This paper tries to address these two fundamental questions presenting an exemplary case study. 

The epistemological disproportion between common sense and irrationality of the creative process became the fundamental basis of the research. It encouraged our interest in reading a generated text as a narrative of madness. Why do we treat machine texts as if they are primitive maxims or well known common knowledge? What if we read them as narratives with the broadest potentiality of meaning like insane notes of asylum patients? Would this approach change the text generation process? 

In this paper, we present the Paranoid Transformer, a fully autonomous text generator that is based on a paranoiac-critical system and aims to change the approach to reading generated texts.  The critical statement of the project is that the absurd mode of reading and the evaluation of generated texts enhances and changes what we understand under computational creativity. Another critical aspect of the project is that Paranoid Transformer resulting text stream is fully unsupervised.
This is a fundamental difference between the Paranoid Transformer and the vast majority of text generation systems presented in the literature that are relying on human post-moderation, i.e., cherry-picking.

Originally, Paranoid Transformer was represented on the National Novel Generation Month contest\footnote{https://github.com/NaNoGenMo/2019} (NaNoGenMo 2019) as an unsupervised text generator that can create narratives in a specific dark style. The project has resulted in a digital mad writer with a highly contextualized personality, which is of crucial importance for the creative process \cite{veale2019Book}. 

\section{Related Work}

There is a variety of works related to the generation of creative texts like the generation of poems, catchy headlines, conversations, and texts in particular literary genres. Here we would like to discuss a certain gap in the field of creative text generation studies and draw attention to the specific reading approach that can lead to more intriguing results in terms of computational creativity. 

The interest in text generation mechanisms is rapidly growing since the arrival of deep learning.  The there are various angles from which researcher approach text generation. For example, \cite{van2019churnalist} and \cite{alnajjar2019no} study generative models that could produce relevant headlines for the news publications. A variety of works study stylization potential of generative models either for prose, see \cite{Jhamtani}, or for poetry, see \cite{tikhonov2018sounds,tikhonov2018guess}. 

Generative poetry dates back as far as \cite{Wheatley} along with other early generative mechanisms and has various subfields at the moment. Generation of poems could be addressed following specific literary tradition, see \cite{Hezhou,Yan2,Yi}; could be focused on the generation of topical poetry \cite{Ghazvininejad}; could be centered around stylization that targets a certain author \cite{Yamshchikov2019}  or a genre \cite{Potash}. For a taxonomy of generative poetry techniques, we address the reader to \cite{Lamb}.

The symbolic notation of music could be regarded as a subfield of text generation, and the research of computational potential in this context has an exceptionally long history. To some extent, it holds a designated place in the computational creativity hall of fame. Indeed, at the very start of computer-science Ada Lovelace already entertains a thought that an analytical engine can produce music on its own.  \cite{menabrea1842sketch} state: "Supposing, for instance, that the fundamental relations of pitched sounds in the science of harmony and of musical composition were susceptible of such expression and adaptations, the engine might compose elaborate and scientific pieces of music of any degree of complexity or extent." For an extensive overview of music generation mechanisms, we address the reader to \cite{briot2019deep}.

One has to mention a separate domain related to different aspects of the 'persona' generation. These could include relatively well-posed problems such as the generation of biographies out of the structured data, see \cite{Lebret}, or open-end tasks for the personalization of dialogue agent, dating back to \cite{weizenbaum1966eliza}. With the rising popularity of chat-bots and the arrival of deep learning, the area of persona-based conversation models \cite{Li2016APN} is growing by leaps and bounds. The democratization of generative conversational methods provided by open-source libraries such as \cite{burtsev2018deeppavlov,shiv2019microsoft} fuels further advancements in this field.

However, the majority of text generation approaches are chasing the generation as the significant value of such algorithms, which makes the very concept of computational creativity seem less critical. Another major challenge is the presentation of the algorithms' output. Vast majority of results on natural language generation either do not imply that generated text has any artistic value, or expect certain post-processing of the text to be done by a human supervisor before the text is presented to the actual reader. We believe that the value of computational creativity is to be restored by shifting the researcher's attention from generation to the process of framing the algorithm \cite{charnley2012notion}. We show that such shift it possible since the generated output of Paranoid Transformer does not need any additional laborious manual post-processing. 

The most reliable framing approaches are dealing with attempts to clarify the algorithm by providing the context, describing the process of generative acts, and making calculations about the generative decisions \cite{cook2019framing}. In this paper, we suggest that such an unusual framing approach as the obfuscation of the produced output could be quite profitable in terms of increasing the number of interpretations and enriching the creative potentiality of generated text.  

Obfuscated interpretation of the algorithm's output methodologically intersects with the literary theory that deals with the reader as the key figure responsible for the meaning. In this context, we aim to overcome disciplinary borderline and create bisociative knowledge, which develops the fundamentals of computational creativity \cite{veale2019computational}. This also goes in line with the ideas of \cite{ohsawa2003modeling,abe2011curation} regarding obfuscation as a mode of reading generated texts that the reader either commits voluntarily or is externally motivated to switch gears and perceive generated text in such mode. This commitment implies a chance discovery of potentially rich associations and extensions of possible meaning.

How exactly can literary theory contribute to computational creativity in terms of the text generation mechanisms? As far as the text generation process implies an incremental interaction between neural networks and a human, it inevitably presupposes critical reading of the generated text. This reading brings a lot in the final result and comprehensibility of artificial writing. In Literature studies, the process of meaning creation is broadly discussed by hermeneutical philosophers, who treated the meaning as a developing relationship between the message and the recipient, whose horizons of expectations are constantly changing and enriching the message with new implications \cite{gadamer1994literature,hirsch1967validity}.

The importance of reception and its difference from author's intentions was convincingly demonstrated and articulated by the so-called Reader-response theory, a particular branch of the Receptive theory that deals with verbalised receptions. As Stanley Fish, one of the principal authors of the approach, puts it, the meaning does not reside in the text but in the mind of the reader \cite{fish1980there}. Thus, any text may be interpreted differently, depending on the reader's background, which means that even an absurd text could be perceived as meaningful under specific circumstances. The same concept was described by \cite{eco1972towards} as so-called aberrant reading and implied that the difference between intention and interpretation is a fundamental principle of cultural communication. It is often the shift in interpretative paradigm that makes remarkable works of art to be dismissed by most at first like Picasso’s Les Demoiselles d’Avignon that was not recognized by artistic society and was not exhibited for nine years since it had been created. 

One of the most recognizable literary abstractions in terms of creative potentiality is the so-called 'romantic mad poet' whose reputation was historically built on the idea that genius would never be understood \cite{whitehead2017madness}. Madness in terms of cultural interpretation is far from its psychiatric meaning and has more in common with the historical concept of a marginalized genius. Mad narrator was chosen as a literary emploi for the Paranoid Transformer to extend the interpretative potentiality of the original text that could be not ideal in formal terms, on the other hand, it could be attributed to an individual with exceptional understanding of the world, which gives more linguistic freedom to this individual for expressing herself and more freedom in interpreting her messages. The anthropomorphization of the algorithm makes the narrative more personal, which is as important as the personality of a recipient in the process of meaning creation \cite{dennett2014self}. The self expression of the Paranoid Transformer is enhanced by introducing a nervous handwriting that amplifies the effect and gives more context for interpretation. In this paper, we show that treating the text generator as a romantic mad poet gives more literary freedom to the algorithm and generally improves the text generation. The philosophical basis of our approach is derived from the idea of creativity as an act of transpassing the borderline between conceptual realms. Thus, the dramatic conflict between computed and creative text could be solved by extending the interpretative horizons. 

\section{Model and Experiments}

The general idea behind the Paranoid Transformer project is to build a 'paranoid' system based on two neural networks. The first network (Paranoid Writer) is a GPT-based \cite{radford2019language} tuned conditional language model, and the second one (Critic subsystem) uses a BERT-based classifier \cite{devlin2019bert} that works as a filtering subsystem. The critic selects the 'best' texts from the generated stream of texts that Paranoid Writer produces and filters the ones that it deems to be useless. Finally, an existing handwriting synthesis neural network implementation is applied to generate a nervous handwritten diary where a degree of shakiness depends on the sentiment strength of a given sentence. This final touch further immerses the reader into the critical process and enhances the personal interaction of the reader with the final text. Shaky handwriting frames the reader and, by design, sends her on the quest for meaning. 

\subsection{Generator Subsystem}

The first network, Paranoid Writer, uses an OpenAI GPT \cite{radford2019language} architecture implementation by huggingface\footnote{https://github.com/huggingface/transformers}. We used a publicly available model that was already pre-trained on a huge fiction BooksCorpus dataset with approximately ~10K books with ~1B words.

The pre-trained model was fine-tuned on several additional handcrafted text corpora, which altogether comprised approximately ~50Mb of text for fine-tuning. These texts included:
\begin{itemize}
    \item a collection of Crypto Texts (Crypto Anarchist Manifesto, Cyphernomicon, etc.);
    \item a collection of fiction books from such cyberpunk authors as Dick, Gibson, and others;
\item non-cyberpunk authors with particular affinity to fringe mental prose, for example, Kafka and Rumi;
    \item  transcripts and subtitles from some cyberpunk movies and series such as Bladerunner;
    \item several thousands of quotes and fortune cookie messages collected from different sources.
\end{itemize}

During the fine-tuning phase, we have used special labels for conditional training of the model:
\begin{itemize}
    \item QUOTE for any short quote or fortune, LONG for others;
    \item CYBER for cyber-themed texts and OTHER for others.
\end{itemize}

 Each text got two labels, for example, it was LONG+CYBER for Cyphernomicon, LONG+OTHER for Kafka, and QUOTE+OTHER for fortune cookie messages. Note, there were almost no texts labeled as QUOTE+CYBER, just a few nerd jokes. The idea of such conditioning and the choice of texts for fine-tuning was rooted in the principle of reading a madness narrative discussed above. The obfuscation principle manifests itself in the fine-tuning on the short aphoristic quotes and ambivalent fortune cookies. It aims to enhance the motivation of the reader and to give her additional interpretative freedom. Instrumentally the choice of the texts was based on two fundamental motivations: we wanted to simulate a particular fringe mental state, and we also were specifically aiming into short diary-like texts to be generated in the end. It is well known that modern state-of-the-art generative models are not able to support longer narratives yet can generate several consecutive sentences that are connected with one general topic. QUOTE/LONG label allowed us to control the model and to target shorter texts during the generation. Such short ambivalent texts could subjectively be more intense. At the same time, inclusion of longer texts in the fine-tuning phase allowed us to shift the vocabulary of the modal even further toward a desirable 'paranoid' state. We also were aiming into some proxy of 'self-reflection' that would be addressed as a topic in the resulting 'diary' of the paranoid transformer. To push the model in this direction, we introduced cyber-themed texts. As a result of these two choices, in generation mode, the model was to generate only QUOTE+CYBER texts. The raw results were already promising enough:

\texttt{let painting melt away every other shred of reason and pain, just lew the paint to move thoughts away from blizzes in death. let it dry out, and turn to cosmic delights, to laugh on the big charms and saxophones and fudatron steames of the sales titanium. we are god's friends, the golden hands on the shoulders of our fears. do you knock my cleaning table over? i snap awake at some dawn. the patrons researching the blues instructor's theories around me, then give me a glass of jim beam. boom!}

However, this was not close enough to any sort of creative process. Our paranoid writer had graphomania too. To amend this mishap and improve the resulting quality of the texts, we wanted to incorporate additional automated filtering.

\subsection{Heuristic Filters}

As a part of the final system, we have implemented heuristic filtering procedures alongside with a critic subsystem. 

The heuristic filters were as follows:
\begin{itemize}
    \item reject the creation of new, non-existing words;
    \item reject phrases with two unconnected verbs in a row;
    \item reject phrases with several duplicating words;
    \item reject phrases with no punctuation or with too many punctuation marks.
\end{itemize}

The application of this script cut the initial text flow into a subsequence of valid chunks filtering the pieces that could not pass the filter. Here are several examples of such chunks after heuristic filtering:

\texttt{a slave has no more say in his language but he has to speak out!}

\texttt{the doll has a variety of languages, so its feelings have to fill up some time of the day - to - day journals. the doll is used only when he remains private. and it is always effective.}

\texttt{leave him with his monk - like body.}

\texttt{a little of technique on can be helpful.}

To further filter the stream of such texts, we implemented a critic subsystem. 

\subsection{Critic subsystem}

We have manually labeled 1 000 of generated chunks with binary labels GOOD/BAD. We marked a chunk as BAD in case it was grammatically incorrect or just too dull or too stupid. The labeling was profoundly subjective. We marked more disturbing and aphoristic chunks as GOOD, pushing the model even further into the desirable fringe state of paranoia simulation. Using these binary labels, we have fine-tuned a pre-trained publicly available BERT classifier\footnote{https://github.com/huggingface/transformers\#model-architectures} to predict the label of any given chunk.

Finally, a pipeline that included the Generator subsystem, the heuristic filters, and the Critic subsystem produced the final results:

\texttt{a sudden feeling of austin lemons, a gentle stab of disgust. i'm what i'm}

\texttt{humans whirl in night and distance.}

\texttt{we shall never suffer this. if the human race came along tomorrow, none of us would be as wise as they already would have been. there is a beginning and an end.}

\texttt{both of our grandparents and brothers are overdue. he either can not agree or he can look for someone to blame for his death.}

\texttt{he has reappeared from the world of revenge, revenge, separation, hatred. he has ceased all who have offended him.}

\texttt{and i don't want the truth. not for an hour.}

The resulting generated texts were already thought-provoking and allowed reading a narrative of madness, but we wanted to enhance this experience and make it more immersive for the reader.

\subsection{Nervous Handwriting}

\begin{figure}[t!]
 \includegraphics[width=0.49\textwidth]{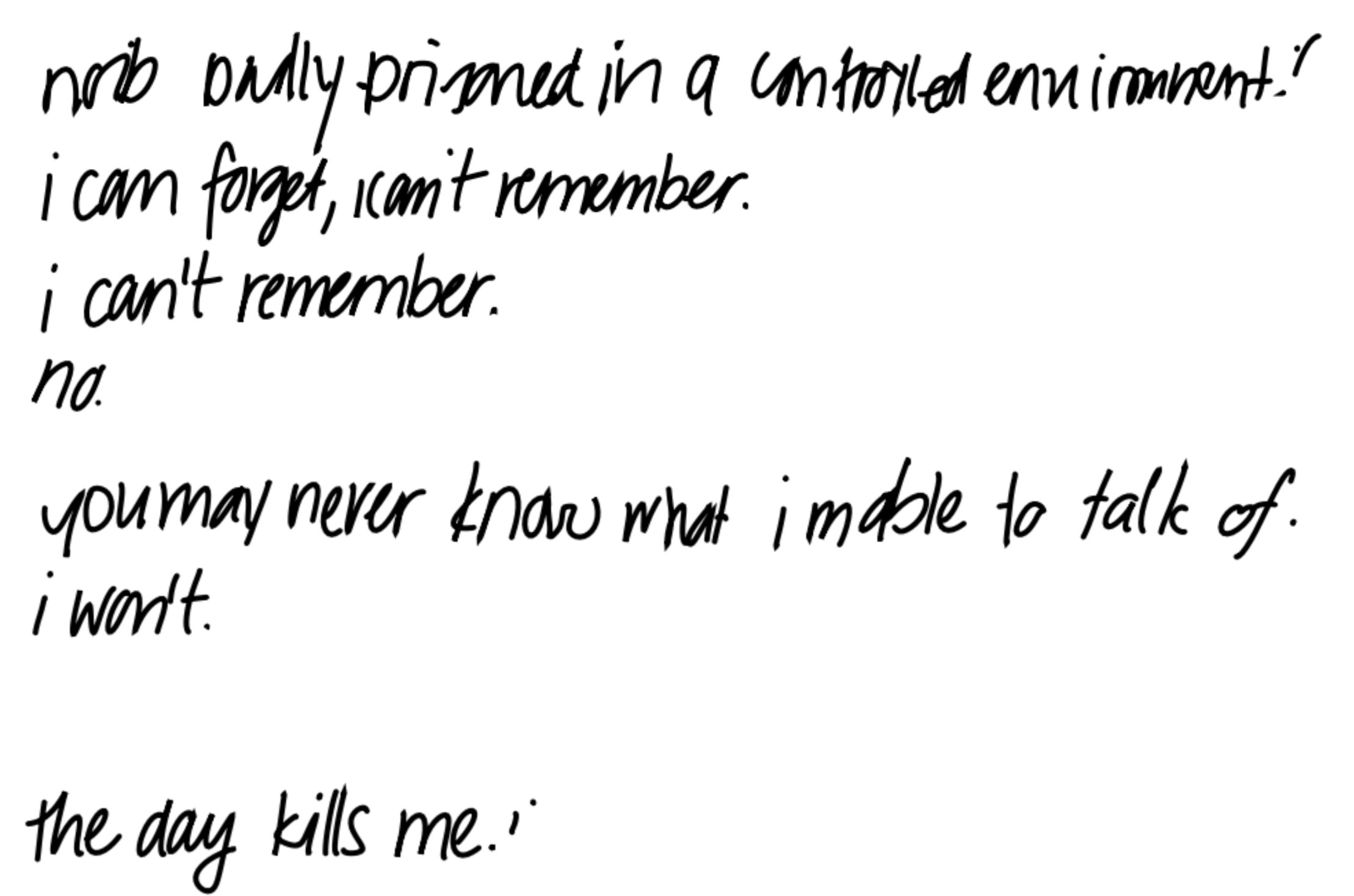}
\caption{
Some examples of Paranoid Transformer diary entries. Three entries of varying length.}
\label{fig:1}    
\end{figure}

\begin{figure}[h!]
 \includegraphics[width=0.5\textwidth]{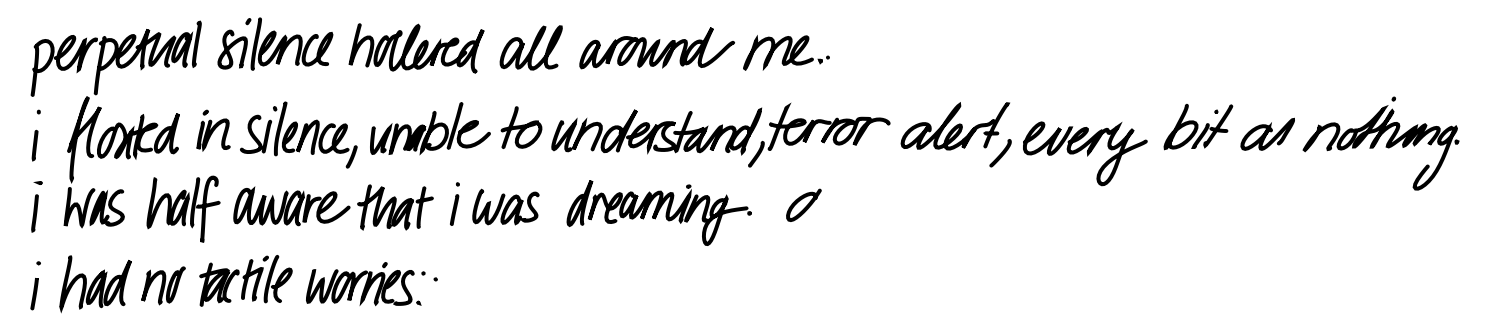}
\caption{
Some examples of Paranoid Transformer diary entries. Longer entry proxying 'self-reflection' and personalized fringe mental state experience.}
\label{fig:2}    
\end{figure}

\begin{figure}[h!]
 \includegraphics[width=0.5\textwidth]{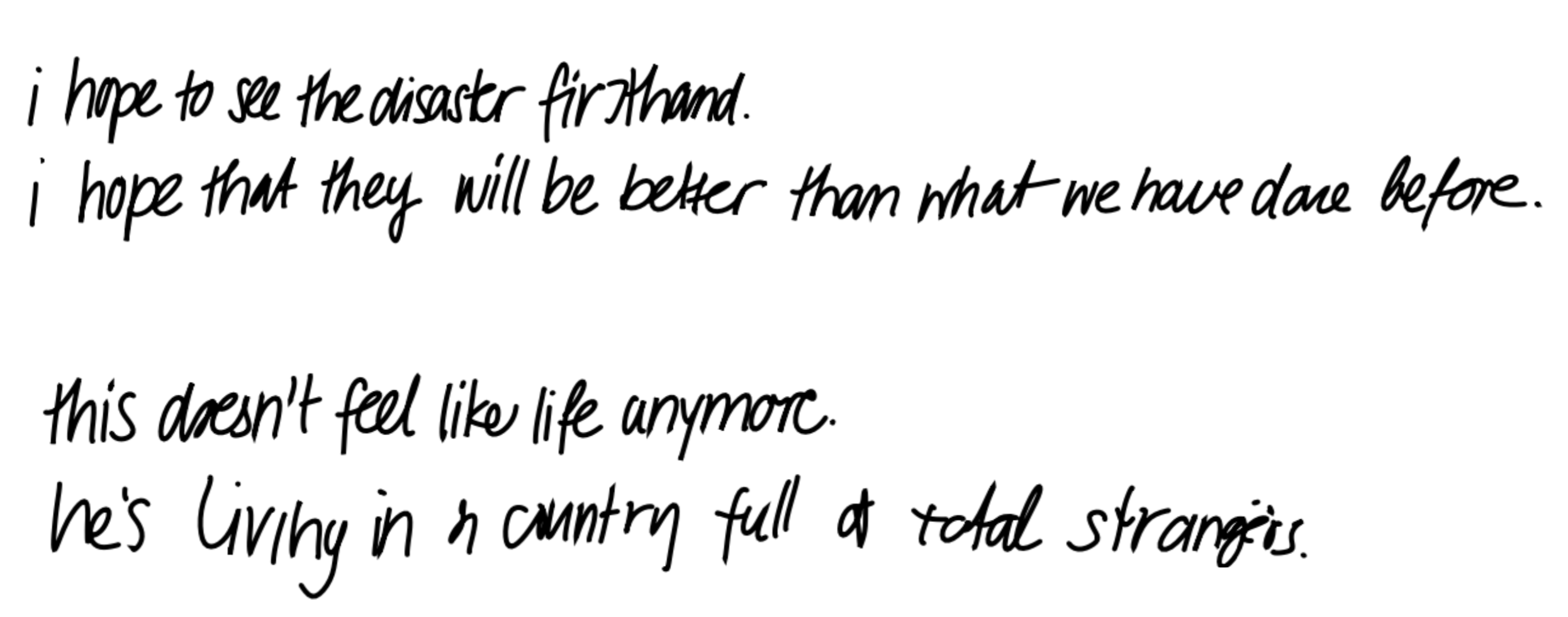}
\caption{
Some examples of Paranoid Transformer diary entries. Typical entries with destructive and ostracised motives.}
\label{fig:3}    
\end{figure}

In order to enhance the personal aspect of the artificial paranoid author, we have implemented an additional generative element. Using implementation\footnote{https://github.com/sjvasquez/handwriting-synthesis} for handwriting synthesis from \cite{Graves}, we have generated handwritten versions of the generated texts. Bias parameter was used to make the handwriting shakier if the generated text's sentiment was stringer. Figures \ref{fig:1}--\ref{fig:3} show several final examples of the Paranoid Transformer diary entries.

Figure \ref{fig:1} demonstrates that the length of the entries can differ from several consecutive sentences that convey a longer line of reasoning to a short, abrupt four-words note.

Figure \ref{fig:2} illustrates typical entry of 'self-reflection'. The text explores the narrative of dream and could be paralleled with a description of an out-of-body experience \cite{blanke2004out} generated by the predominantly out-of-body entity.

Figure \ref{fig:3} illustrates typical entries with destructive and ostracised motives. This is an exciting side-result of the model that we did not expect. The motive of loneliness is recurring in the Paranoid Transformer diaries. 

It is important to emphasize that the resulting stream of the generated output is available online\footnote{https://github.com/altsoph/paranoid\_transformer}. No human post-processing of output is performed.

\section{Discussion}

In Dostoevsky’s “Notes from the Underground” there is a striking idea about madness as a source of creativity and computational explanation as a killer of artistic magic: “We sometimes choose absolute nonsense because in our foolishness we see in that nonsense the easiest means for attaining a supposed advantage. But when all that is explained and worked out on paper (which is perfectly possible, for it is contemptible and senseless to suppose that some laws of nature man will never understand), then certainly so-called desires will no longer exist." \cite{dostoevsky1984zapiski} Paranoid Transformer brings forward an important question about the limitations of the computational approach of creative intelligence, either it belongs to a human or algorithm. This case demonstrates that creative potentiality and generation efficiency could be considerably influenced by such poorly controlled methods as obfuscated supervision and loose interpretation of the generated text.

Creative text generation studies inevitably strive to reveal fundamental cognitive structures that can explain the creative thinking of a human. The suggested framing approach to machine narrative as a narrative of madness brings forward some crucial questions about the nature of creativity and the research perspective on it. In this section, we are going to discuss the notion of creativity that emerges from the results of our studying and reflect on the framing of the text generation algorithm.

What does creativity in terms of text generation mean? Is it a cognitive production of novelty or rather generation of unexpendable meaning? Can we identify any difference in treating human and machine creativity? 

In his groundbreaking work \cite{turing1950computing} pinpoints several crucial aspects of intelligence. He states: "If the meaning of the words “machine” and “think” are to be found by examining how they are commonly used it is difficult to escape the conclusion that the meaning and the answer to the question, “Can machines think?” is to be sought in a statistical survey such as a Gallup poll." This starting argument turned out to be prophetic. It pinpoints the profound challenge for the generative models that use statistical learning principles. Indeed, if creativity is something on the fringe, on the tails of the distribution of outcomes, then it is hard to expect a model that is fitted on the center of distribution to behave in a way that could be subjectively perceived as a creative one. Paranoid Transformer is a result of a conscious attempt to push the model towards a fringe state of proximal madness. This case study serves as a clear illustration that creativity is onthologically opposed to the results of the "Gallup poll."

\begin{figure}[h!]
 \includegraphics[width=0.45\textwidth]{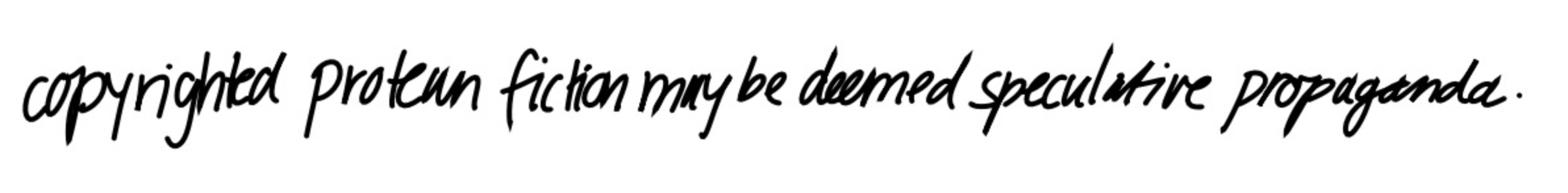}
\caption{
"Copyrighted protein fiction may be deemed speculative propaganda," -- the authors are tempted to proclaim this diary entry the motto of Paranoid Transformer.}
\label{fig:4}    
\end{figure}

Another question that raises discussion around computational creativity deals with a highly speculative notion of self within a generative algorithm. Does a mechanical writer have a notion of self-expression? Considering a wide range of theories of the self (carefully summarized in \cite{jamwal2019}), a creative AI generator triggers a new philosophical perspective on this question. As any human self, an artificial self does not develop independently. By following John Locke’s understanding of self as based on memory  \cite{locke1860essay}, Paranoid Transformer builds itself on memorising the interactive experience with a human, furthermore, it emotionally inherits to its supervising readers who labelled the training dataset of the supervision system. On the other hand, Figure \ref{fig:4} clearly shows the impact of crypto-anarchic philosophy on the Paranoid Transformers' notion of self. One can easily interpret the paranoiac utterance of the generator as a doubt about reading and processing unbiased literature.

According to the cognitive science approach, the construction of self could be revealed in narratives about particular aspects of self \cite{dennett2014self}. In the case of Paranoid Transformer, both visual and verbal self-representation result in nervous and mad narratives that are further enhanced by the reader.

Regarding the problem of framing the study on creative text generators, we cannot avoid the question concerning the novelty of the generated results. Does Paranoid Transformer demonstrate a new result that is different from others in the context of computational creativity? First of all, we can use external validation. At the moment, the Paranoid Transformer' book of is prepared to come out of print. Secondly, and probably more importantly here, we can indicate the novelty of the conceptual framing of the study. Since the design and conceptual situatedness influence the novelty of the study \cite{perivsic2019situated}, we claim that the suggested conceptual extension of perceptive horizons of interaction with generative algorithm can solely advocate the novelty of the result. 

An important question that deals with framing of the text generation results engages the discussion about the possibility of a chance discovery. In \cite{ohsawa2003modeling} lays out three crucial three keys for chance discovery, namely, communication, context shifting, and data mining.  \cite{abe2011curation} further enhances these ideas addressing the issue of curation and claiming that a curation is a form of communication. The Paranoid Transformer is a clear case study that is rooted in Ohsawa's three aspects of chance discovery. Data mining is represented with a choice of data for fine-tuning and the process of fine-tuning itself. Communication is interpreted under Abe's broader notion of curation as a form of communication. Context shift manifests itself thought the reading the narrative of madness that invests the reader with interpretative freedom and motivates her to pursue the meaning in her own mind though simple, immersive visualization of the systems' fringe 'mental state'.

\section{Conclusion}

This paper presents a case study of a Paranoid Transformer. It claims that framing the machine-generated narrative as a narrative of madness can intensify the personal experience of the reader. We explicitly address three critical aspects of chance discovery and claim that the resulting system could be perceived as a digital persona in a fringe mental state. The crucial aspect of this perception is the reader, who is motivated to invest meaning into the resulting generative texts. This motivation is built upon several pillars: a challenging visual form, that focuses the reader on the text; obfuscation, that opens the resulting text to broader interpretations; and the implicit narrative of madness, that is achieved with the curation of the dataset for the fine-tuning of the model. Thus we intersect the understanding of computational creativity with the fundamental ideas of receptive theory.








\bibliographystyle{iccc}
\bibliography{iccc}

\end{document}